# Bound Propagation

**Martijn Leisink**                                    MARTIJN@MBFYS.KUN.NL
**Bert Kappen**                                         BERT@MBFYS.KUN.NL
*University of Nijmegen, Department of Biophysics*
*Geert Grooteplein 21, 6525EZ Nijmegen, The Netherlands*

## Abstract

In this article we present an algorithm to compute bounds on the marginals of a graphical model. For several small clusters of nodes upper and lower bounds on the marginal values are computed independently of the rest of the network. The range of allowed probability distributions over the surrounding nodes is restricted using earlier computed bounds. As we will show, this can be considered as a set of constraints in a linear programming problem of which the objective function is the marginal probability of the center nodes. In this way knowledge about the maginals of neighbouring clusters is passed to other clusters thereby tightening the bounds on their marginals. We show that sharp bounds can be obtained for undirected and directed graphs that are used for practical applications, but for which exact computations are infeasible.

## 1. Introduction

Graphical models have become a popular and powerful tool to deal with dependencies in a complex probability model. For small networks an exact computation of the marginal probabilities for certain nodes is feasible (see, for example Zhang & Poole, 1994). Unfortunately, due to an exponential scaling of the computational complexity, these exact algorithms soon fail for somewhat more complex and more realistic networks.

To deal with this computational problem two main streams can be distinguished. Firstly, there is the approach of approximating the exact solution. Numerous algorithms have been developed among which are mean field (Peterson & Anderson, 1987; Saul et al., 1996) and TAP (Thouless et al., 1977; Plefka, 1981). Nowadays, approximation methods such as belief propagation (Murphy et al., 1999; Yedidia et al., 2000) using the Bethe or Kikuchi energy are gaining popularity. In contrast with mean field expansions, the latter methods try to model only the local probability distributions of several small clusters, which communicate with each other in a way known as message passing.

The second approach is that of finding upper and lower bounds on important quantities of the graphical model. In contrast to approximation techniques, these methods give reliable information. One may find, for instance, that a marginal probability is definitely between 0.4 and 0.5. These methods usually focus on the partition function of the network, which can be lower bounded (Leisink & Kappen, 2001) or upper bounded (Wainwright et al., 2002). Little work has been done, however, on bounding the marginal probabilities for certain nodes in the network directly. Up to now researchers have focused their attention on combining upper and lower bounds on the partition function in which way bounds on the marginals can be derived. For the special case of the Boltzmann machine this was done by Leisink and Kappen (2002).

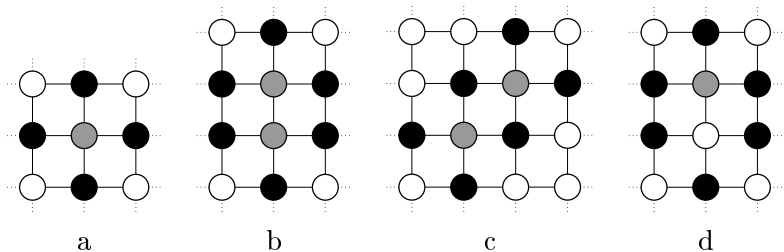

Figure 1: Examples of separator and marginal nodes in an (infinite) Ising grid. Light shaded nodes correspond to $S_{\mathrm{mar}}$, black nodes to $S_{\mathrm{sep}}$. (a) A single node enclosed by its tightest separator. (b), (c) Two marginal nodes enclosed by their tightest separator. (d) Another choice for a separator for a single node marginal; the other node enclosed by that separator belongs to $S_{\mathrm{oth}}$.

In this article we present a method called 'bound propagation' which has certain similarities to belief propagation in the sense that small clusters of a few nodes try to bound their marginals as tight as possible. The knowledge contained in these bounds is shared with surrounding clusters. Iteration of this procedure improves bounds on all marginals until it is converged. It is important to realize that bounding and approximation techniques are not mutually exclusive. We propose to always run the bound propagation algorithm first. When the result is not satisfactory, one can always fall back upon approximation techniques as belief propagation paying the price of being uncertain about the solution.

In Section 2 we explain the general method of bound propagation, which is described more algorithmicly in Section 3. In Section 4 we show the behaviour of the algorithm on a toy problem, whereas in Section 5 the results are shown for more realistic networks. Finally, in Section 6, we conclude and discuss possible paths for future research.

## 2. What Bounds Can Learn From Each Other

Consider a Markov network defined by the potentials $\psi_i (S_i)$ each dependent on a set of nodes $S_i$. The probability to find the network in a certain state $S$ is proportional to

$$p(S) \propto \prod_i \psi_i (S_i) \tag{1}$$

Problems arise when one is interested in the marginal probability over a small number of nodes (denoted by $S_{\mathrm{mar}}$), since in general this requires the summation over all of the exponentially many states.

Let us define the set of separator nodes, $S_{\mathrm{sep}}$, to be the nodes that separate $S_{\mathrm{mar}}$ from the rest of the network. One can take, for instance, the union of the nodes in all potentials that contain at least one marginal node, while excluding the marginal nodes itself in that union (which is the Markov blanket of $S_{\mathrm{mar}}$):

$$S_{\mathrm{sep}} = \bigcup_{S_i \cap S_{\mathrm{mar}} \neq \emptyset} S_i / S_{\mathrm{mar}} \tag{2}$$

See for instance Figure 1a–c. It is not necessary, however, that $S_{\text{sep}}$ defines a tight separation. There can be a small number of other nodes, $S_{\text{oth}}$, inside the separated area, but not included by $S_{\text{mar}}$ (Figure 1d). A sufficient condition for the rest of the theory is that the size of the total state space, $\|S_{\text{sep}} \cup S_{\text{mar}} \cup S_{\text{oth}}\|$, is small enough to do exact computations.

Suppose that we know a particular setting of the separator nodes, $\vec{s}_{\text{sep}}$, then computing the conditional distribution of $S_{\text{mar}}$ given $S_{\text{sep}}$ is easy:

$$p\left(\vec{s}_{\text{mar}} \mid \vec{s}_{\text{sep}}\right) = \sum_{\vec{s}_{\text{oth}} \in S_{\text{oth}}} p\left(\vec{s}_{\text{mar}}\vec{s}_{\text{oth}} \mid \vec{s}_{\text{sep}}\right) \tag{3}$$

This expression is tractable to compute since $S_{\text{sep}}$ serves as the Markov blanket for the union of $S_{\text{mar}}$ and $S_{\text{oth}}$. Thus the actual numbers are independent of the rest of the network and can be found either by explicitly doing the summation or by any other more sophisticated method. Similarly, if we would know the exact distribution over the separator nodes, we can easily compute the marginal probability of $S_{\text{mar}}$:

$$p\left(\vec{s}_{\text{mar}}\right) = \sum_{\vec{s}_{\text{sep}} \in S_{\text{sep}}} \sum_{\vec{s}_{\text{oth}} \in S_{\text{oth}}} p\left(\vec{s}_{\text{mar}}\vec{s}_{\text{oth}} \mid \vec{s}_{\text{sep}}\right) p\left(\vec{s}_{\text{sep}}\right) \tag{4}$$

Unfortunately, in general we do not know the distribution $p\left(\vec{s}_{\text{sep}}\right)$ and therefore we cannot compute $p\left(\vec{s}_{\text{mar}}\right)$. We can, however, still derive certain properties of the marginal distribution, namely upper and lower bounds for each state of $S_{\text{mar}}$. For instance, an upper bound on $\vec{s}_{\text{mar}}$ can easily be found by computing

$$p_{+}\left(\vec{s}_{\text{mar}}\right) = \max_{q\left(\vec{s}_{\text{sep}}\right)} \sum_{\vec{s}_{\text{sep}} \in S_{\text{sep}}} \sum_{\vec{s}_{\text{oth}} \in S_{\text{oth}}} p\left(\vec{s}_{\text{mar}}\vec{s}_{\text{oth}} \mid \vec{s}_{\text{sep}}\right) q\left(\vec{s}_{\text{sep}}\right) \tag{5}$$

under the constraints $\forall_{\vec{s}_{\text{sep}} \in S_{\text{sep}}}\ q\left(\vec{s}_{\text{sep}}\right) \geq 0$ and $\sum_{\vec{s}_{\text{sep}} \in S_{\text{sep}}} q\left(\vec{s}_{\text{sep}}\right) = 1$. The distribution $q\left(\vec{s}_{\text{sep}}\right)$ corresponds to $\|S_{\text{sep}}\|$ free parameters (under the given constraints) used to compute the worst case (here: the maximum) marginal value for each $\vec{s}_{\text{mar}}$. Obviously, if $q\left(\vec{s}_{\text{sep}}\right) = p\left(\vec{s}_{\text{sep}}\right)$ the upper bound corresponds to the exact value, but in general $p\left(\vec{s}_{\text{sep}}\right)$ is unknown and we have to maximize over all possible distributions. One may recognize Equation 5 as a simple linear programming (LP) problem with $\|S_{\text{sep}}\|$ variables. With a similar equation we can find lower bounds $p_{-}\left(\vec{s}_{\text{mar}}\right)$.

Upto now we did not include any information about $q\left(\vec{s}_{\text{sep}}\right)$ we might have into the equation above. But suppose that we already computed upper and lower bounds on the marginals of other nodes than $S_{\text{mar}}$. How can we make use of this knowledge? We investigate our table of all sets of marginal nodes we bounded earlier and take out those that are subsets of $S_{\text{sep}}$. Obviously, whenever we find already computed bounds on $S'_{\text{mar}}$ where $S'_{\text{mar}} \subseteq S_{\text{sep}}$[1], we can be sure that

$$\sum_{\vec{s}_{\text{sep}} \in S_{\text{sep}}/S'_{\text{mar}}} q\left(\vec{s}_{\text{sep}}\right) = q\left(\vec{s}'_{\text{mar}}\right) \leq p_{+}\left(\vec{s}'_{\text{mar}}\right) \tag{6}$$

and similarly for the lower bound. These equations are nothing more than extra contraints on $q\left(\vec{s}_{\text{sep}}\right)$ which we may include in the LP-problem in Equation 5. There can be several $S'_{\text{mar}}$ that are subsets of $S_{\text{sep}}$ each defining $2\|S'_{\text{mar}}\|$ extra constraints.

---

1. In the rest of this article we will use the symbol $S'_{\text{mar}}$ to indicate an earlier computed bound over $S'_{\text{mar}}$ with $S'_{\text{mar}} \subseteq S_{\text{sep}}$.

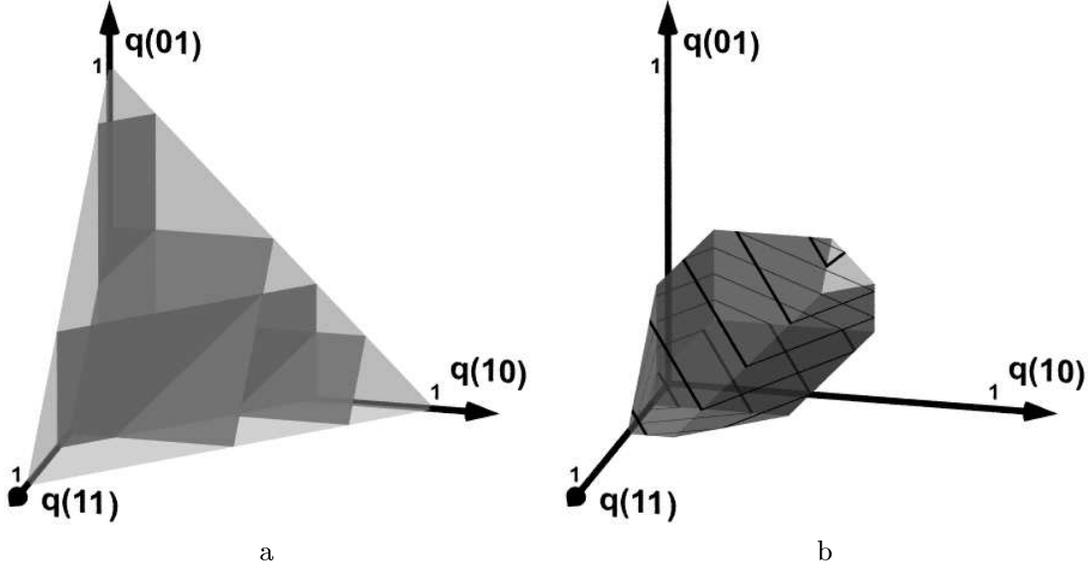

Figure 2: The area in which $q(s_1 s_2)$ must lie is shown. $q(00)$ is implicitly given since the distribution is normalized. a) The pyramid is the allowable space. The darker planes show how this pyramid can be restricted by adding earlier computed bounds as constraints in the linear programming problem. b) This results in a smaller polyhedron. The black lines show the planes where the function $\sum_{s_1 s_2} p(s_0 = 1 \mid s_1 s_2) q(s_1 s_2)$ is constant for this particular problem.

In this way, information collected by bounds computed earlier, can propagate to $S_{\text{mar}}$, thereby finding tighter bounds for these marginals. This process can be repeated for all sets of marginal nodes we have selected until convergence is reached.

## 2.1 Linear Programming

In terms of standard linear programming, Equation 5 can be expressed as

$$\max \left\{ \vec{c} \cdot \vec{x} \ \middle| \ \vec{x} \geq 0 \ \wedge \ A\vec{x} \leq \vec{b} \right\} \tag{7}$$

where the variables are defined as

$$\vec{x} = q(\vec{s}_{\text{sep}}) \tag{8}$$

$$\vec{c} = \sum_{\vec{s}_{\text{oth}} \in S_{\text{oth}}} p(\vec{s}_{\text{mar}} \vec{s}_{\text{oth}} \mid \vec{s}_{\text{sep}}) = p(\vec{s}_{\text{mar}} \mid \vec{s}_{\text{sep}}) \tag{9}$$

$$A = \left\{ \begin{array}{l} \delta(\vec{s}'_{\text{mar}}, \vec{s}_{\text{sep}}) \\ -\delta(\vec{s}'_{\text{mar}}, \vec{s}_{\text{sep}}) \end{array} \right. \qquad \vec{b} = \left\{ \begin{array}{l} p_+(\vec{s}'_{\text{mar}}) \\ -p_-(\vec{s}'_{\text{mar}}) \end{array} \right. \qquad \left( \forall_{S'_{\text{mar}} \subseteq S_{\text{sep}}} \forall_{\vec{s}'_{\text{mar}} \in S'_{\text{mar}}} \right) \tag{10}$$

where $\delta(\vec{s}'_{\text{mar}}, \vec{s}_{\text{sep}}) = 1$ iff the states of the nodes both node sets have in common are equal. Thus the columns of the matrix $A$ correspond to $\vec{s}_{\text{sep}}$, the rows of $A$ and $b$ to all the constraints (of which we have $2 \, \|S'_{\text{mar}}\|$ for each $S'_{\text{mar}} \subseteq S_{\text{sep}}$). The ordering of the rows of $A$ and $b$ is irrelevant as long as it is the same for both. The constraint that $q(\vec{s}_{\text{sep}})$ should be normalized can be incorporated in $A$ and $\vec{b}$ by requiring

$$\sum_{\vec{s}_{\text{sep}}} q(\vec{s}_{\text{sep}}) \leq 1 \qquad \text{and} \qquad -\sum_{\vec{s}_{\text{sep}}} q(\vec{s}_{\text{sep}}) \leq -1 \qquad (11)$$

The maximum of $\vec{c} \cdot \vec{x}$ corresponds to $p_+(\vec{s}_{\text{mar}})$. The negative of $p_-(\vec{s}_{\text{mar}})$ can be found by using $-\vec{c} \cdot \vec{x}$ as the objective function.

## 2.2 A Simple Example

Imagine a single neuron, $s_0$, that is separated from the rest of the network by two other neurons, $s_1$ and $s_2$. We want to compute bounds on the marginal $p(s_0 = 1)$. Since we do not know the distribution over the separator nodes, we have to consider all possible distributions as in Equation 5. Therefore we introduce the free parameters $q(s_1 s_2)$. The goal is now to compute the minimum and maximum of the function $\sum_{s_1 s_2} p(s_0 = 1 \mid s_1 s_2) q(s_1 s_2)$. In Figure 2a we show in three dimensions the space (the large pyramid) in which the distribution $q(s_1 s_2)$ lies. $q(00)$ is implicitly given by one minus the three other values.

We can add, however, the earlier computed (single node) bounds on $p(s_1)$ and $p(s_2)$ to the problem. These restrict the space in Figure 2a further, since for instance (see also Equation 6)

$$q(s_1 = 1) = \sum_{s_2} q(s_1 = 1, s_2) = q(10) + q(11) \leq p_+(s_1 = 1) \qquad (12)$$

We have four independent constraints, which are shown in Figure 2a as planes in the pyramid.

Obviously, by adding this information the space in which $q(s_1 s_2)$ may lie is restricted to that shown in Figure 2b. In the same figure we have added black lines, which correspond to the planes where the objective function is constant. A standard linear programming tool will immediately return the maximum and the minimum of this function thus bounding the marginal $p(s_0 = 1)$.

## 3. The Algorithm In Detail

To solve an arbitrary network we first determine the set $\Omega$ of all $S_{\text{sep}}$ we will use for the problem. We choose

$$\Omega(N) = \left\{ S_{\text{sep}} \mid \|S_{\text{cl}}\| \leq N \wedge \forall_{s \notin S_{\text{cl}}} \|S_{\text{cl}} \cup s\| > N \right\} \qquad (13)$$

where $S_{\text{cl}}$ is the cluster $S_{\text{sep}} \cup S_{\text{mar}} \cup S_{\text{oth}}$ and $s$ denotes a single node. This choice limits the state space of $S_{\text{cl}}$ to $N$, which more or less determines the computation time. If we choose, for example, $N = 2^5$ for the Ising grid in Figure 1 assuming binary nodes, only the configuration in Figure 1a would be included in $\Omega$. Choosing $N = 2^8$ puts the configurations in Figure 1b–d and a few others not shown into $\Omega$, but not Figure 1a, since that separator is already embedded in larger ones (e.g. Figure 1d). In other words: $\Omega(N)$ is the set of all $S_{\text{sep}}$ for which $\|S_{\text{sep}} \cup S_{\text{mar}} \cup S_{\text{oth}}\|$ is not larger than $N$, but if we add a single node it would cross that boundary.

We will compute bounds for all $S_{\mathrm{mar}}$ for which there can be found a separator in $\Omega$. We reserve memory for $p_+ \left( S_{\mathrm{mar}} \right)$ and $p_- \left( S_{\mathrm{mar}} \right)$ and initialize them to one and zero respectively. Note that if we have a bound over $S_{\mathrm{mar}}$ it is still worthwhile to compute bounds over subsets of $S_{\mathrm{mar}}$ (e.g. Figure 1b and d). In contrast to a joint probability table, the bounds on marginals over subsets cannot be computed simply by summing over the marginal of $S_{\mathrm{mar}}$, since we have only bounds and not the real values.

## 3.1 The Iterative Part

For all $S_{\mathrm{sep}} \in \Omega$ we perform the same action. First we set up the $A$ and $\vec{b}$ for the LP-problem, since this depends only on $S_{\mathrm{sep}}$ and the already computed bounds for which $S'_{\mathrm{mar}} \subseteq S_{\mathrm{sep}}$. The number of variables in our LP-problem obviously is $\| S_{\mathrm{sep}} \|$. The number of (inequality) constraints is equal to

$$\sum_{S'_{\mathrm{mar}} \subseteq S_{\mathrm{sep}}} 2 \left\| S'_{\mathrm{mar}} \right\| \tag{14}$$

Once the LP-problem is set up this far, we try to improve all bounds that are defined over an $S_{\mathrm{mar}}$ for which $S_{\mathrm{sep}}$ is its separator. The separator in figures 1b and d, for instance, is identical, but the $S_{\mathrm{mar}}$'s differ. For each bound we iterate over its state space and set the $\vec{c}$ as in Equation 9 to its appropriate value. Then we compute the new upper and lower bounds for that $\vec{s}_{\mathrm{mar}}$ by solving the LP-problem twice: maximize and minimize. If the new found value improves the bound, we store it, otherwise we abandon it.

The iterative procedure is repeated until convergence is reached. In our simulations we define a bound as being converged as soon as the relative improvement after one iteration is less than one percent. If this holds for all bounds, the algorithms stops.

## 3.2 Computational Complexity

LP-problems have been thoroughly studied in the literature. The computational complexity of such problems is shown to be polynomial (Khachiyan, 1979). But more importantly, for most practical problems the number of iterations needed is close to linear in the number of variables or the number of constraints, whichever is less (Todd, 2002). In each iteration a 'pivoting action' needs to be performed, which is some operation that roughly accesses all the elements of the matrix $A$ once. Therefore the expected computational complexity for solving one LP-problem is $\mathcal{O} \left( m^2 \| S_{\mathrm{sep}} \| \right)$, where $m$ is the number of constraints. The actual observed computation time depends heavily on the particular problem, but in general LP-problems upto tens of thousands of variables and constraints are reasonable. This makes the presented method tractable for separators with $\| S_{\mathrm{sep}} \|$ of a similar size.

For every $\vec{s}_{\mathrm{mar}} \in S_{\mathrm{mar}}$ we need to solve two LP-problems: one for the upper and one for the lower bound. Note that in this iteration, we do not need to change the matrix $A$ and vector $\vec{b}$ from Equation 10. The vector $\vec{c}$, that defines the objective function, obviously does vary with $\vec{s}_{\mathrm{mar}}$. Therefore, we expect a total computational complexity of $\mathcal{O} \left( m^2 \| S_{\mathrm{mar}} \| \| S_{\mathrm{sep}} \| \right)$ for updating one cluster $S_{\mathrm{mar}}$. How quickly the algorithm converges is more difficult to estimate. In the next section, however, we address this topic on the basis of a toy problem.

To conclude this section, we remark that when the LP-problem is not tractable, one can always leave out as many constraints as needed, paying the price of getting looser bounds.

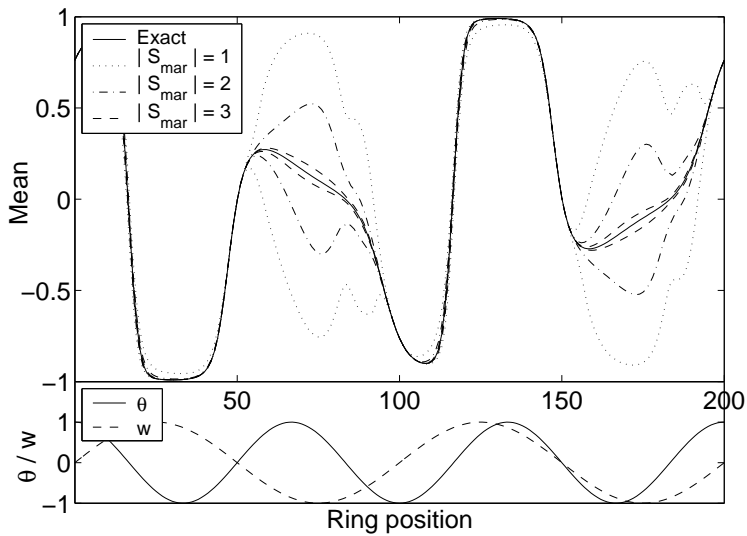

Figure 3: A Boltzmann machine ring of 200 nodes is initialized with thresholds and weights as shown in the lower part of the figure. Exact means are plotted as well as the results of the bound propagation algorithm. The number of nodes included in $S_{\mathrm{mar}}$ is indicated in the legend.

## 4. A Toy Problem

To get some intuition about the performance of the algorithm we created a ring (one large loop) of 200 binary nodes. Nodes are connected as in the Boltzmann machine with a weight $w$ and each node has a threshold $\theta$. Weights and thresholds are varying along the ring as is shown at the bottom of Figure 3. For this simple problem we can compute the exact single node marginals. These are plotted as a solid line in Figure 3. We ran the bound propagation algorithm three times, where we varied the maximum number of nodes included in $S_{\mathrm{mar}}$: $|S_{\mathrm{mar}}| = 1$, $|S_{\mathrm{mar}}| = 2$ and $|S_{\mathrm{mar}}| = 3$. In all cases we chose two separator nodes: the two neighbors of $S_{\mathrm{mar}}$. Thus these three cases correspond to $\Omega\left(2^3\right)$, $\Omega\left(2^4\right)$ and $\Omega\left(2^5\right)$. As is clear from Figure 3 the simplest choice already find excellent bounds for the majority of the nodes. For large negative weights, however, this choice is not sufficient. Allowing larger separators clearly improves the result. In all cases the computation time was a few seconds[2].

### 4.1 Fixed Points and Convergence

It is possible to derive the fixed point of the bound propagation algorithm analytically for a Boltzmann ring if we take a single value for all weights $(w)$ and one for all thresholds $(\theta)$. Due to this symmetry all single node marginals are equal in such a network. Moreover all upper and lower bounds on the single nodes should be identical. This implies that for the

_______________
2. We used a computer with a 1 GHz Athlon processor.

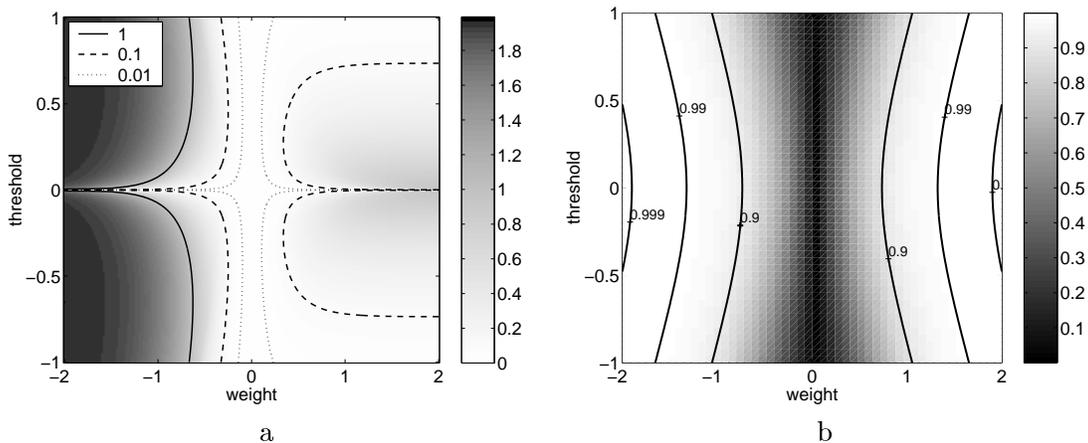

Figure 4: The result for a Boltzmann machine ring of an arbitrary number of nodes. All thresholds and weights have the same value. In the left panel the gap is shown between the upper and lower bound of the mean of the nodes. The algorithm converges exponentially to the final bounding values as soon as it is close enough. Thus the distance to its asymptotic value is proportional to $\alpha^n$, where $n$ is the number of iterations. The parameter $\alpha$ is shown in the right panel.

fixed point the following holds for any $i$:

$$p_+ (s_i) = \max_{q(s_{i-1}s_{i+1})} p(s_i|s_{i-1}s_{i+1}) q(s_{i-1}s_{i+1}) \tag{15}$$

under the constraints

$$q(s_{i-1}) \leq p_+ (s_{i-1}) = p_+ (s_i) \tag{16}$$

$$q(s_{i+1}) \leq p_+ (s_{i+1}) = p_+ (s_i) \tag{17}$$

$$q(s_{i-1}) \geq p_- (s_{i-1}) = p_- (s_i) \tag{18}$$

$$q(s_{i+1}) \geq p_- (s_{i+1}) = p_- (s_i) \tag{19}$$

and similarly for the lower bound. The conditional probability in Equation 15 is given by

$$p(s_i|s_{i-1}s_{i+1}) \propto \exp(\theta s_i + w s_{i-1}s_i + w s_i s_{i+1}) \tag{20}$$

From these equations one can derive the fixed point for $p_+ (s_i)$ and $p_- (s_i)$. It turns out, however, that it is easier to determine the fixed point of the upper and lower bound on the mean, hence $p_+ (s_i = 1) - p_- (s_i = -1)$ and $p_- (s_i = 1) - p_+ (s_i = -1)$. But this has no effect on the principle results as tightness and convergence speed.

In Figure 4a the difference between the upper and lower bound on the means is shown for various choices of the weight and threshold. As we saw before tight bounds can be obtained for small weights and somewhat larger, but positive weights, whereas negative weights result in rather poor, but still non-trivial bounds. It should be mentioned, however, that these results correspond to the choice of the smallest clusters ($|S_{\mathrm{mar}}| = 1$ and $|S_{\mathrm{sep}}| = 2$) for the bound propagation algorithm. The bounds can easily be improved by enlarging the clusters as we saw in Figure 3.

Close to the fixed point the bound propagation algorithm converges exponentially. The distance to the asymptotic value can be written as $\alpha^n$, where $n$ is the number of iterations and $\alpha$ is a number between zero and one indicating the convergence speed. The closer to one $\alpha$ is, the slower the algorithm converges. In Figure 4b $\alpha$ is shown for the same weights and thresholds. It is clear that larger weights induce a slower convergence. That is what we see in general: probabilities that are close to deterministic slow down the convergence.

## 5. Real World Networks

To assess the quality of the bound propagation algorithm for more challenging networks, we computed bounds for a Bayesian network (Section 5.1), that is commonly used in articles as a representative of a real world problem. Secondly, we show that the bound propagation method can be used to find many tight bounds for a large Ising grid (Section 5.2 and finally, in Section 5.3, we show results for a bi-partite graph. For these networks, we will briefly discuss bound propagation in relation with the cluster variation method (Kappen & Wiegerinck, 2002) which is one of the better approximation methods for general graphs. Whenever possible, we show the exact marginals.

### 5.1 The Alarm Network

The Alarm network[3] is a commonly used network which is a representitive of a real life Bayesian network. It was originally described by Beinlich et al. (1989) as a network for monitoring patients in intensive care. It consists of 37 nodes of two, three or four states and is small enough to do exact computations. In Figure 5 the graphical model is shown. Our task was to determine the single node marginals in absence of any evidence. For each node the exact marginal probabilities for its states are shown as horizontal lines. The top and bottom of the rectangles around these lines indicate the upper and lower bounds respectively. For this network we used $\Omega(25,000)$[4]. If we make this choice, we can treat exactly the minimal Markov blanket of all single node marginals. The algorithm converged in about six minutes.

We see that the bounds in the lower left corner are very close to the exact value. We can give some intuition why this happens. Observe that node 7 on its own can play the role of separator for nodes 1 to 6. Thus all correlations between these nodes and the rest of the network are through node 7, which means that their uncertainty only depends on the uncertainty of node 7. The latter happens to be very small (consider that as a given fact) and therefore the marginal bounds for nodes 1 to 6 are very tight.

The bounds in the upper right corner, on the other hand, are quite poor. This is partially due to the density of the connections there and the number of states each of the nodes has. In combination this leads to large state spaces already for small clusters. Therefore, we cannot compute bounds on sets of multiple nodes, which generally leads to weaker bounds.

---

3. The data can be downloaded from the Bayesian Network Repository at
   `http://www.cs.huji.ac.il/labs/compbio/Repository/`
4. We started with $\Omega(100)$ and gradually increased this value up to 25,000. The bounds obtained with small clusters are used as good starting points for the more time consuming larger clusters.

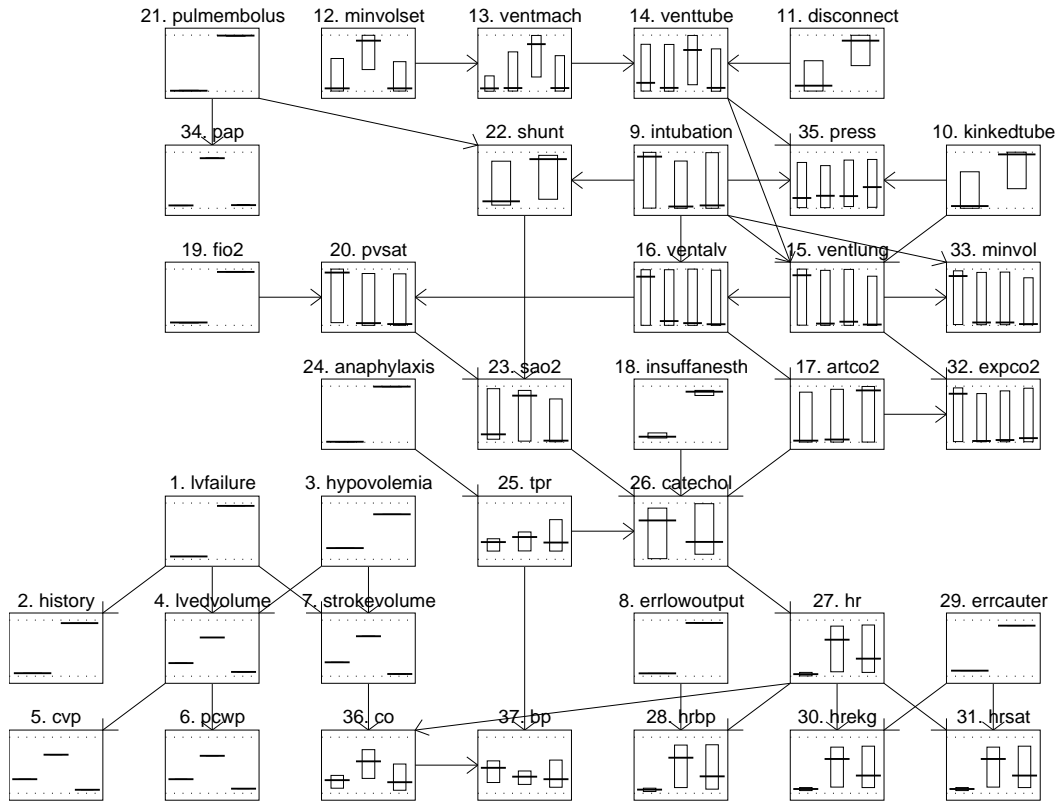

Figure 5: The exact marginals (thick horizontal lines) and the bounds (rectangles) for the alarm network. Some bounds are so tight that the rectangle is not visible. We used $\Omega(25,000)$.

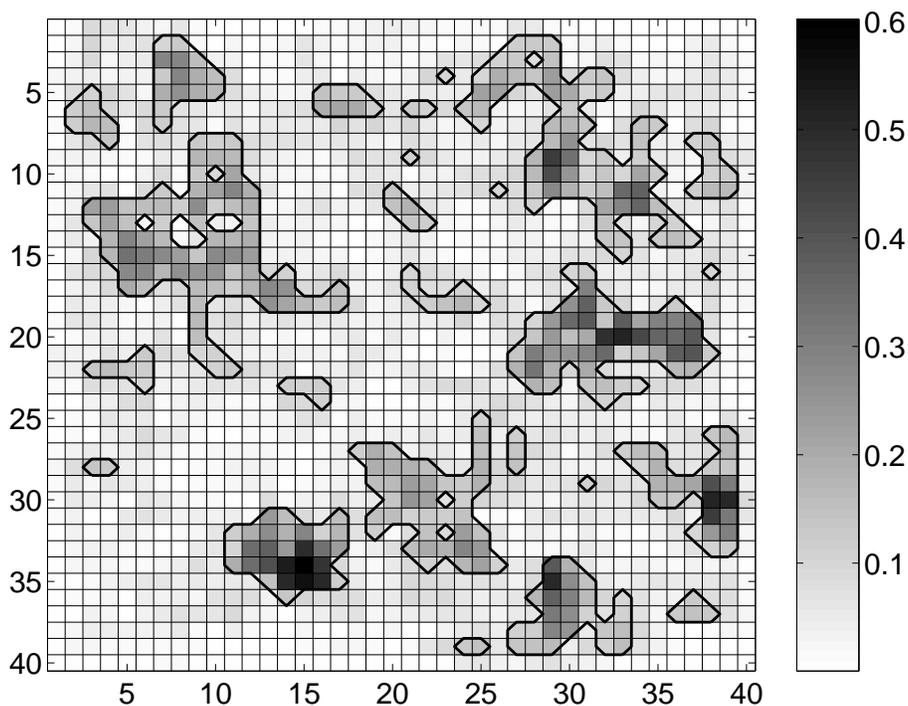

Figure 6: The band width (upper minus lower bound) of each node in a 40x40 Ising grid is shown by the blackness of the squares. The exact marginal could not be computed for such a large network. The solid lines show the boundary where the band width was 0.1. We used $\Omega$ (2500).

Secondly, the local probability tables are of a form that is difficult for the bound propagation algorithm. That is, a lot of probabilities are close to zero and one.

This picture is what we generally see. There are some parts in the network that are hard to compute (not necessarily impossible) which usually have a more dense structure and a larger (local) state space. Other parts can be bounded very well. Note that this network could easily be embedded in a very large network that is intractable for any exact method. Even in that case, the bound propagation algorithm could be used to find similar results. Since the method is based on local computations, the existance of a large surrounding network has almost no influence.

## 5.2 A Large Ising Grid

We created a so called two-dimensional Ising model, which is a rectangular grid with connections between all pairs of nodes that are direct neighbours. We used a grid of 40 by 40 binary nodes. The potentials were drawn from a uniform distribution between zero and one. In contrast with Bayesian belief networks (Neal, 1992) the probability distribution over the unclamped Ising grid is not automatically normalized, but this has no consequences for the algorithm.

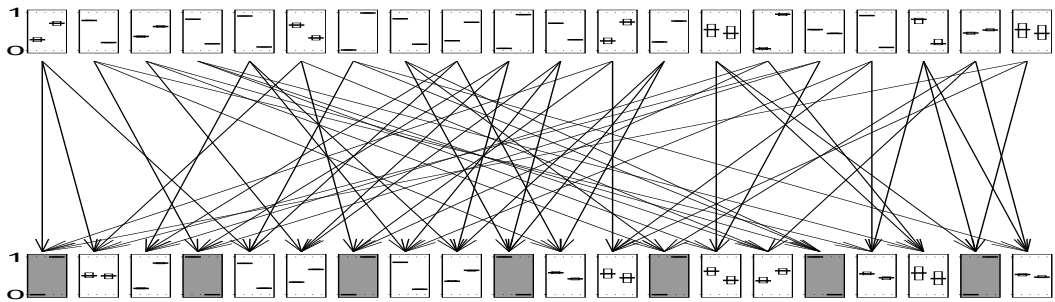

Figure 7: A bi-partite network of twenty nodes in each layer. Each child node (bottom layer) has three parents in the top layer and each parent node has exactly three children. Probability tables were randomly initialized from a uniform distribution. The shaded nodes in the bottom layer are clamped to a certain state. The bound propagation algorithm ran with $\Omega\,(2500)$ and converged in 37 seconds. For each node the thick horizontal lines show their exact marginals, the top and bottom of the rectangle are the upper and lower bounds found.

For small Ising grids computing the exact marginals is tractable, but as soon as the grid is larger than about 25 by 25 the exact algorithm fails due to the exponential scaling of the computational complexity. The bound propagation algorithm, on the other hand, only depends on the local structure (i.e. the size of the Markov blankets of each node) and thus scales linearly with the number of nodes. We created an 40x40 Ising grid with binary nodes similarly as above and ran the bound propagation algorithm. For this network the exact algorithm would require a storage capacity of at least $2^{41}$ real numbers, whereas bound propagation converged in about 71 minutes in which time it computed bounds on the marginals over all 1600 nodes.

In Figure 6 we show the 40x40 grid where the blackness of the squares correspond to the band width of the single node marginals. This band width is defined as the difference between the upper and lower bound. Due to the fact that marginal probabilities sum to one, the two band widths for these binary neurons are identical. We can clearly see some regions in the lattice (the blacker area's) for which bound propagation had some difficulties. Most of the marginals, however, are bounded quite well: more than 75% had a band width less than 0.1.

## 5.3 The Bi-Partite Graph

A bi-partite graph is a network consisting of two layers of nodes, where the top layer is hidden and the bottom layer visible. The only connections in the network are from the top layer (parent nodes) to the bottom layer (child nodes). Such a architecture appears very simple, but already with several tens of nodes it is often intractable to compute the marginals exactly when evidence is presented. A bi-partite graph can be undirected or directed (pointing downwards). A recent example of the first is the POE (product of experts) from Hinton (1999) with discrete nodes. The 'Quick Medical Reference' (QMR) network from Shwe et al. (1991) is a good example for a directed bi-partite graph. Recently there were some approaches to use approximation techniques for this network (Jaakkola & Jordan, 1999; Murphy et al., 1999). Therefore we have chosen to use the bi-partite network as an architecture to test the bound propagation algorithm on.

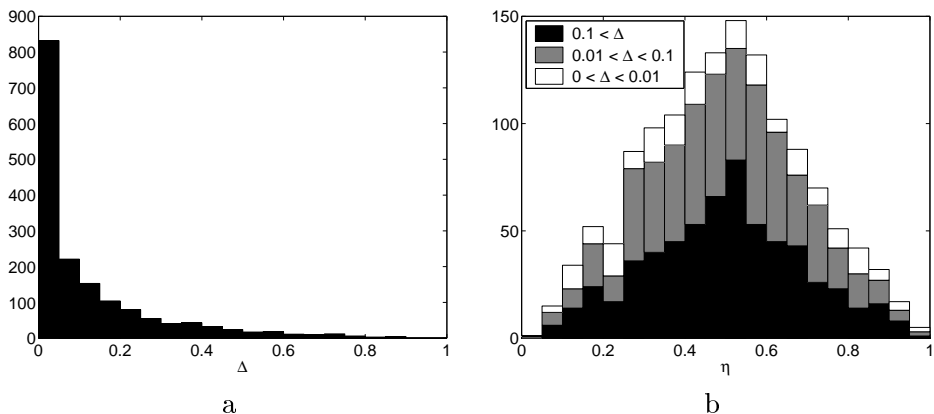

Figure 8: A bi-partite network of a thousand nodes in each layer was created and evidence was presented on every third child node. The left histogram shows the band widths found for the single node marginals of all unclamped nodes. The bound propagation algorithm converged in about 75 minutes and used $\Omega(2500)$. The right histogram shows the relative position of the approximation within the bounding region. All approximations were between the upper and lower bound, although this is not necessarily the case.

We created bi-partite networks of an equal number of (binary valued) nodes in both layers. Each child node has exactly three parents and each parent points to exactly three children. The connections were made randomly. The conditional probability tables (CPT's) holding the probability of the child given its parents was initialized with uniform random number between zero and one. The prior probability tables for the parents were initialized similarly.

In Figure 7 we show a bi-partite network with twenty nodes in each layer, which is still small enough to compute the exact marginals. The arrows show the parent-child relations. To every third child evidence was presented (the shaded nodes). We ran the bound propagation algorithm with $\Omega(2500)$ and reached convergence in 37 seconds. For every node we show the marginal over its two states. The thick line is the exact marginal, the top and bottom of the rectangle around it indicate the upper and lower bound respectively. It is clear that for most of the nodes the computed bounds are so tight that they give almost the exact value. For every node (except two) the gap between upper and lower bound is less than one third of the whole region.

The network shown in Figure 7 is small enough to be treated exactly. This enables us to show the computed bounds together with the exact marginals. The bound propagation algorithm, however, can be used for much larger architectures. Therefore we created a bi-partite graph as we did before, but this time with a thousand nodes in each layer. Again we presented evidence to every third node in the bottom layer. The bound propagation algorithm using $\Omega$ (2500) converged in about 75 minutes. In Figure 8a we show a histograms of the band widths (upper minus lower bound) found for the single node marginals for all unclamped nodes. Clearly, for the majority of the marginals very tight bounds are found.

Although exact computations are infeasible for this network, one can still use approximation methods to find an estimate of the single node marginals. Therefore, we ran the cluster variation method (CVM) as described in (Kappen & Wiegerinck, 2002). For the clusters needed by CVM, we simply took all node sets for which a potential is defined. For every unclamped node we computed the relative position of the approximation for the single node marginal within the bounding interval, denoted by $\eta$. Thus $\eta = 0$ indicates an approximation equal to the lower bound, and similarly $\eta = 1$ for the upper bound. Although there is no obvious reason the approximated marginals should be inside the bounding interval, it turned out they were. A histogram of all the relative positions is shown in Figure 8b, where we split up the full histogram into three parts, each referring to a certain region of band widths. It is remarkable that the shape of the histogram does not vary much given the tightness of the bounds. Approximations tend to be in the middle, but for all band widths there are approximations close to the edge.

One can argue that we are comparing apples to pears, since we can easily improve the results of both algorithms, while it is not clear which cases are comparable. This is certainly true, but it is not our intention to make a comparison to determine which method is the best. Approximation and bounding methods both have their own benefits. We presented Figure 8b to give at least some intuition about the actual numbers. In general we found that approximations are usually within the bounding intervals, when computation time is kept roughly the same. This does, however, not make the bounds irrelevant. On the contrary, one could use the bounding values as a validation whether approximations are good or even use them as confidence intervals.

## 6. Discussion

We have shown that bound propagation is a simple algorithm with surprisingly good results. It performs exact computations on local parts of the network and keeps track of the uncertainties that brings along. In this way it is capable to compute upper and lower bounds on any marginal probability of a small set of nodes in the intractable network. Currently we do not understand which network properties are responsible for the tightness of the bounds found. In Figure 6, for instance, we saw islands of worse results in a sea of tight bounds. It is obvious that one bad bound influences its direct neighbourhood, since bound propagation performs local computations in the network. We have no clue, however, why these islands are at the location we found them. We tried to find a correlation with the size of the weights (rewriting the network potentials to a Boltzmann distribution) or with the local frustration of the network (spin glass state), but could not explain the quality of the bounds in terms of these quantities. Here we pose it as an open question.

The computational complexity of the algorithm is mainly determined by the state space of the separator nodes one uses, which makes the algorithm unsuitable for heavily connected networks such as Boltzmann machines. Nevertheless, there is a large range of architectures for which bound propagation can easily be done. Such networks typically have a limited number of connections per node which makes the majority of the Markov blankets small enough to be tractable for the bound propagation algorithm.

We want to discuss one important property of the bound propagation algorithm we did not address before. In this article we found our set of $S_{\mathrm{sep}}$'s by Equation 13. Due to this choice the separator nodes will always be in the neighbourhood of $S_{\mathrm{mar}}$. We have, however, much more freedom to choose $S_{\mathrm{sep}}$. In Section 2 we stated that a sufficient condition to make the algorithm tractable is that $\|S_{\mathrm{sep}} \cup S_{\mathrm{mar}} \cup S_{\mathrm{oth}}\|$ is small enough to do exact computations. A more general, but still sufficient condition is that we should choose $S_{\mathrm{sep}}$ such that $p(S_{\mathrm{mar}} \mid S_{\mathrm{sep}})$ can be computed efficiently, since this is the quantity we need (see Equation 9). If the network structure inside the area separated by $S_{\mathrm{sep}}$ can be written as a junction tree with a small enough tree width, we can compute these conditional probabilities even if the number of nodes enclosed by $S_{\mathrm{sep}}$ is very large. For certain architectures the separator can even in that case be small enough. One can think of a network consisting of a number of tractable junction trees that are connected to each other by a small number of nodes. One should be aware of the important difference between the method outlined here and the method of conditioning (Pearl, 1988), which does exact inference. We do not require the part of the network that is outside the separator (i.e. all nodes not in $S_{\mathrm{sep}}$, $S_{\mathrm{mar}}$ or $S_{\mathrm{oth}}$) to be tractable. The open problem is to develop an efficient algorithm to find suitable separators, since the nodes in such an $S_{\mathrm{sep}}$ are generally spread widely.

To conclude this article we like to say a few more words about linear programming. This is a field of research that is still developing. Any improvements on LP-solving algorithms directly influence the results presented in this article. We could imagine that more advanced LP-methods could benefit from the fact that the matrix $A$ in Equation 10 is sparse. At least half of the entries are zero. To be more precise: the constraints induced by a particular $S'_{\mathrm{mar}}$ are exactly $2\|S'_{\mathrm{mar}}\|$ rows in the matrix $A$ with exactly $2\|S_{\mathrm{sep}}\|$ non-zero entries, thus having a fraction of $1 - 1/\|S'_{\mathrm{mar}}\|$ zeros. Moreover, all the non-zero entries are ones. Finally, there is a promising future for LP-solvers, since the algorithms seems to be suitable for parallel computing (Alon & Megiddo, 1990).

## Acknowledgments

This research is supported by the Technology Foundation STW, applied science devision of NWO and the technology programme of the Ministry of Economic Affairs.